\documentclass[runningheads]{llncs}
\usepackage{graphicx}

\usepackage{tikz}
\usepackage{comment}
\usepackage{amsmath,amssymb} 
\usepackage{color}

\usepackage[accsupp]{axessibility}  

\usepackage{kotex}
\usepackage{array}
\usepackage{multirow}
\usepackage{booktabs}
\usepackage{multirow}
\usepackage{multicol}
\usepackage{color}
\usepackage[breaklinks=true]{hyperref}
\usepackage{breakcites}

\newcolumntype{C}{>{\centering\arraybackslash}p{5em}}
\newcolumntype{D}{>{\centering\arraybackslash}p{4em}}
\newcolumntype{E}{>{\centering\arraybackslash}p{7em}}
\newcolumntype{B}{>{\centering\arraybackslash}p{15em}}
\newcolumntype{L}{>{\raggedright\arraybackslash}p{15em}}

\begin{document}
\pagestyle{headings}
\mainmatter
\def\ECCVSubNumber{4076}  

\title{Global-local Motion Transformer for Unsupervised Skeleton-based Action Learning}

    \titlerunning{Global-local Motion Transformer}
    %
    \author{Boeun Kim\inst{1,2} \and
    Hyung Jin Chang\inst{3} \and
    Jungho Kim\inst{2} \and
    Jin Young Choi\inst{1}}
    \authorrunning{Kim et al.}
    %
    \institute{ASRI, Dept. of ECE., Seoul National University \and
    AIRC, Korea Electronics Technology Institute \and
    School of Computer Science, University of Birmingham
    }
\maketitle

\begin{abstract}
We propose a new transformer model for the task of unsupervised learning of skeleton motion sequences.
The existing transformer model utilized for unsupervised skeleton-based action learning is learned the instantaneous velocity of each joint from adjacent frames without global motion information.
Thus, the model has difficulties in learning the attention globally over whole-body motions and temporally distant joints.
In addition, person-to-person interactions have not been considered in the model.
To tackle the learning of whole-body motion, long-range temporal dynamics, and person-to-person interactions, we design a global and local attention mechanism, where, global body motions and local joint motions pay attention to each other.
In addition, we propose a novel pretraining strategy, multi-interval pose displacement prediction, to learn both global and local attention in diverse time ranges.
The proposed model successfully learns local dynamics of the joints and captures global context from the motion sequences.
Our model outperforms state-of-the-art models by notable margins in the representative benchmarks.
Codes are available at \href{https://github.com/Boeun-Kim/GL-Transformer}{\textit{\color{magenta}https://github.com/Boeun-Kim/GL-Transformer}}.
\end{abstract}

\section{Introduction}
In skeleton-based action recognition, to avoid expensive and time-consuming annotation for supervised learning, recent studies have focused on unsupervised learning techniques for pretraining ~\cite{zheng2018unsupervised,su2020predict,lin2020ms2l,xu2021prototypical,kundu2019unsupervised,nie2020unsupervised,rao2021augmented,wang2021contrast,su2021self,cheng2021hierarchical}.
For unsupervised pretraining suitable for action recognition, learning the global context of the entire motion sequence is essential along with learning local joint dynamics and topology.
However, existing methods have limitations in effectively capturing both global context and local joint dynamics.

Several existing unsupervised pretraining methods exploit RNN-based encoder-decoder models~\cite{zheng2018unsupervised,su2020predict,lin2020ms2l,xu2021prototypical,kundu2019unsupervised,nie2020unsupervised}.
However, RNN-based methods have difficulties in extracting global contexts because of the long-range dependency problem~\cite{cho2020self,cai2020learning}.
Other approaches utilize contrastive learning schemes~\cite{rao2021augmented,wang2021contrast,li20213d}. 
However, the performance of these methods has been reported to be highly dependent on the selection of the encoder model because the contrastive loss does not induce detailed learning in the local dynamics of the joints~\cite{wang2021contrast,li20213d}.

Recently, the transformer, widely used for natural language processing and image recognition, has been applied to the unsupervised pretraining of skeleton-based action recognition. The first and the only model is H-transformer~\cite{cheng2021hierarchical}, which learns to predict the direction of the instantaneous velocity of joints in each frame.
H-transformer still has limitations in learning global attention because predicting only the instantaneous velocity induces the model to learn the local attention rather than the global context in whole-body motions.
In addition, H-transformer does not consider person-to-person interactions which are important for classifying actions performed by two or more persons.

In this paper, to tackle the learning of global context, long-range temporal dynamics, and person-to-person interactions, we propose a novel transformer-based pretraining model, which is called GL-Transformer.
To this end, we design the GL-Transformer architecture that contains global and local attention (GLA) mechanism.
The GLA mechanism comprises spatial multi-head attention (spatial-MHA) and temporal multi-head attention (temporal-MHA) modules.
Using the input body motions disentangled into global body motions and local joint motions, 
the spatial-MHA module  performs three types of attention: local(inter-joint), global(body)-from/to-local(joint), and global(person)-to-global(person) attentions.
The temporal-MHA module performs global and local attention between any two frames for sequences of every person.

In addition, a novel pretraining strategy is proposed to induce GL-Transformer to learn global attention across the long-range sequence. For the pretraining, we design a multi-task learning strategy referred to as multi-interval pose displacement prediction (MPDP).
For MPDP, GL-Transformer is trained with multiple tasks to predict multiple pose displacements (angle and movement distance of every joint) over different intervals at the same time. GL-Transformer learns local attention from a small interval, as well as global attention from a large interval.
To enhance performance, we add two factors to GL-Transformer.
First, to learn natural joint dynamics across frames, we impose natural-speed motion sequences instead of sequences sampled to a fixed length.
Next, we introduce a trainable spatial-temporal positional embedding and inject it to each GL-Transformer block repeatedly to use the order information in every block, which is the valuable information of the motion sequence.

We demonstrate the effectiveness of our method through extensive experimental evaluations on widely used datasets: NTU-60~\cite{shahroudy2016ntu}, NTU-120~\cite{liu2019ntu}, and NW-UCLA~\cite{wang2014cross}.
In the linear evaluation protocol~\cite{zheng2018unsupervised}, the performance of GL-Transformer exceeds that of H-transformer~\cite{cheng2021hierarchical} and other state-of-the-art (SOTA) methods by notable margins. 
Furthermore, our method even outperforms SOTA methods in semi-supervised settings. 
The main contributions of this study are summarized as follows:
\begin{enumerate}
    \item We design a novel transformer architecture including global and local attention (GLA) mechanism to model local joint dynamics and capture the global context from skeleton motion sequences with multiple persons (Sec.~\ref{sec:model architecture}).
    \item We introduce a novel pretraining strategy, multi-interval displacement prediction (MPDP), to learn attention in diverse temporal ranges (Sec.~\ref{sec:pretraining_strategy}).
    \item GL-Transformer renews the state-of-the-art score in extensive experiments on three representative benchmarks: NTU-60, NTU-120, and NW-UCLA.
\end{enumerate}

\section{Related Works}

\subsubsection{Unsupervised Skeleton-based Action Recognition.}
Earlier unsupervised learning methods for skeleton-based action recognition can be divided into two categories: using RNN-based encoder-decoders and contrastive learning schemes.
Several existing methods utilize RNN-based encoder-decoder networks~\cite{zheng2018unsupervised,su2020predict,lin2020ms2l,xu2021prototypical,kundu2019unsupervised,nie2020unsupervised}.
The decoder of these networks performs a pretraining task to induce the encoder to extract an appropriate representation for action recognition.
The decoder of LongT GAN~\cite{zheng2018unsupervised} reconstructs the randomly corrupted input sequence conditioned on the representation.
MS$^2$L~\cite{lin2020ms2l} learns to generate more general representations through multi-task learning, which performs tasks such as motion prediction and jigsaw puzzle recognition.
Recently, Colorization~\cite{yang2021skeleton} adopted a GCN to pretrain which regresses the temporal and spatial orders of a skeleton sequence.
RNN-based models suffer from long-range dependencies, and the GCN-based models have a similar challenge because they deliver information sequentially along a fixed path~\cite{plizzari2021spatial,cho2020self,cai2020learning}.
Therefore, the RNN and GCN-based methods have limitations in extracting global representations from the motion sequence, especially from long motions.

Other methods exploit the contrastive learning scheme~\cite{rao2021augmented,wang2021contrast,su2021self}.
These methods augment the original motion sequence and regard it as a positive sample while considering other motion sequences as negative samples.
The model is then trained to generate similar representations between the positive samples using contrastive loss.
AS-CAL~\cite{rao2021augmented} leverages various augmentation schemes such as rotation, shear, reverse, and masking.
Contrastive learning schemes have a limitation, in that all sequences other than themselves are regarded as negative samples, even sequences belonging to the same class.
CrosSCLR~\cite{li20213d} alleviates this issue by increasing the number of positive samples using representations learned from other views, such as velocity or bone sequences.
Because the contrastive learning loss adjusts the distances between the final representations extracted by an encoder, it is difficult to train the encoder to reflect the local joint dynamics explicitly by the loss.
To address the limitations in both categories of unsupervised action recognition, we introduce the transformer~\cite{vaswani2017attention} architecture for modeling the local dynamics of joints and capturing the global context from motion sequences.

\subsubsection{Transformer-based Supervised Learning.}
Transformer-based models have achieved remarkable success in various supervised learning tasks using motion sequences, owing to their attention mechanism, which is suitable for handling long-range sequences.
In supervised action recognition tasks, recent transformer-based methods~\cite{plizzari2021spatial,cho2020self,bai2021gcst,mazzia2022action} outperform GCN-based methods, which have limitations in yielding rich representations because of the fixed graph topology of the human body. 
In the motion prediction task, the method in~\cite{cai2020learning,aksan2021spatio} employs a transformer encoder to capture the spatial-temporal dependency of a given motion sequence and a transformer decoder to generate future motion sequences.
In the 3D pose estimation task, the method in~\cite{zheng20213d} imposes a 2D pose sequence on the spatial-temporal transformer to model joint relations and estimate the 3D pose of the center frame accurately. 

\subsubsection{Transformer-based Pretraining.}
Transformer-based pretraining has become the dominant approach in natural language processing~\cite{devlin2018bert,liu2019roberta}, and is being actively introduced to other research fields such as vision-language~\cite{lu2019vilbert,sun2019videobert,huang2020pixel}, images~\cite{dosovitskiy2020image,he2021masked,chen2021pre,bao2021beit}, and videos~\cite{wang2021bevt,liu2021video}.
The H-transformer~\cite{cheng2021hierarchical} is the first transformer-based pretraining method for motion sequences.
The proposed pretraining strategy predicts the direction of the instantaneous velocity of the joints in each frame. 
This strategy focuses on learning attention from adjacent frames rather than from distant frames.
The model is designed to learn spatial attention between five body part features, where global body movement is not considered.
To address these limitations, we propose a GL-Transformer that contains a global and local attention mechanism and a novel pretraining strategy.
We aim to train GL-Transformer to generate a representation of input motion sequences suitable for the downstream action recognition task by modeling local and global attention effectively in the pretraining process.

\section{Proposed Method}

\subsection{Overall Scheme}
\begin{figure*}[b!]
\centering
    \includegraphics[width=11.5cm]{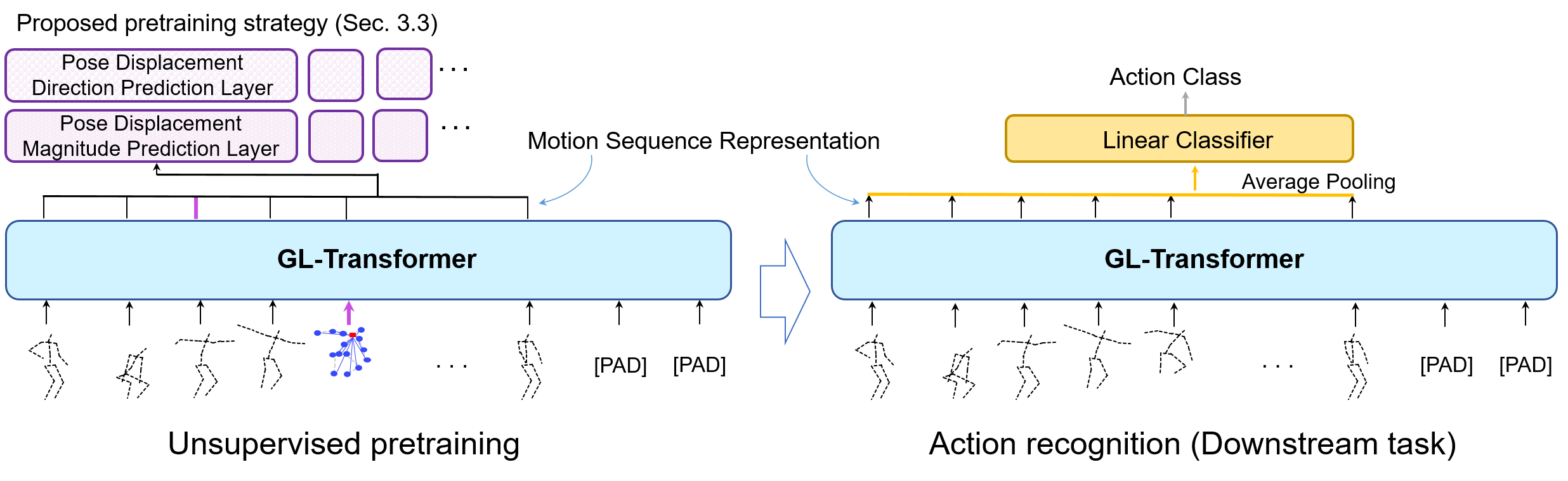}
    \caption{Overall scheme of the proposed framework. GL-Transformer is pretrained with unlabeled motion sequences, and then evaluated in downstream action recognition task}
    
    \label{fig:framework}
\end{figure*}
Our goal is to build a transformer architecture suitable for the skeleton motion sequence (Sec.~\ref{sec:model architecture}) and design a novel pretraining strategy (Sec.~\ref{sec:pretraining_strategy}) for encoding both the internal dynamics and the global context of the motion sequence.
As illustrated in Fig.~\ref{fig:framework}, the proposed framework comprises two stages: unsupervised pretraining and downstream action recognition stages.
In the first stage, we pretrain the proposed transformer-based model, GL-Transformer, with unlabeled motion sequences.
Next, we verify that GL-Transformer generates the appropriate motion representation required for the action recognition.
A single linear classifier is attached after GL-Transformer.
After average pooling is applied to the motion sequence representation for the temporal axis, it is passed to the classifier.

\subsection{Model Architecture}
\label{sec:model architecture}
\begin{figure*}[t]
\centering
    \includegraphics[width=12.0cm]{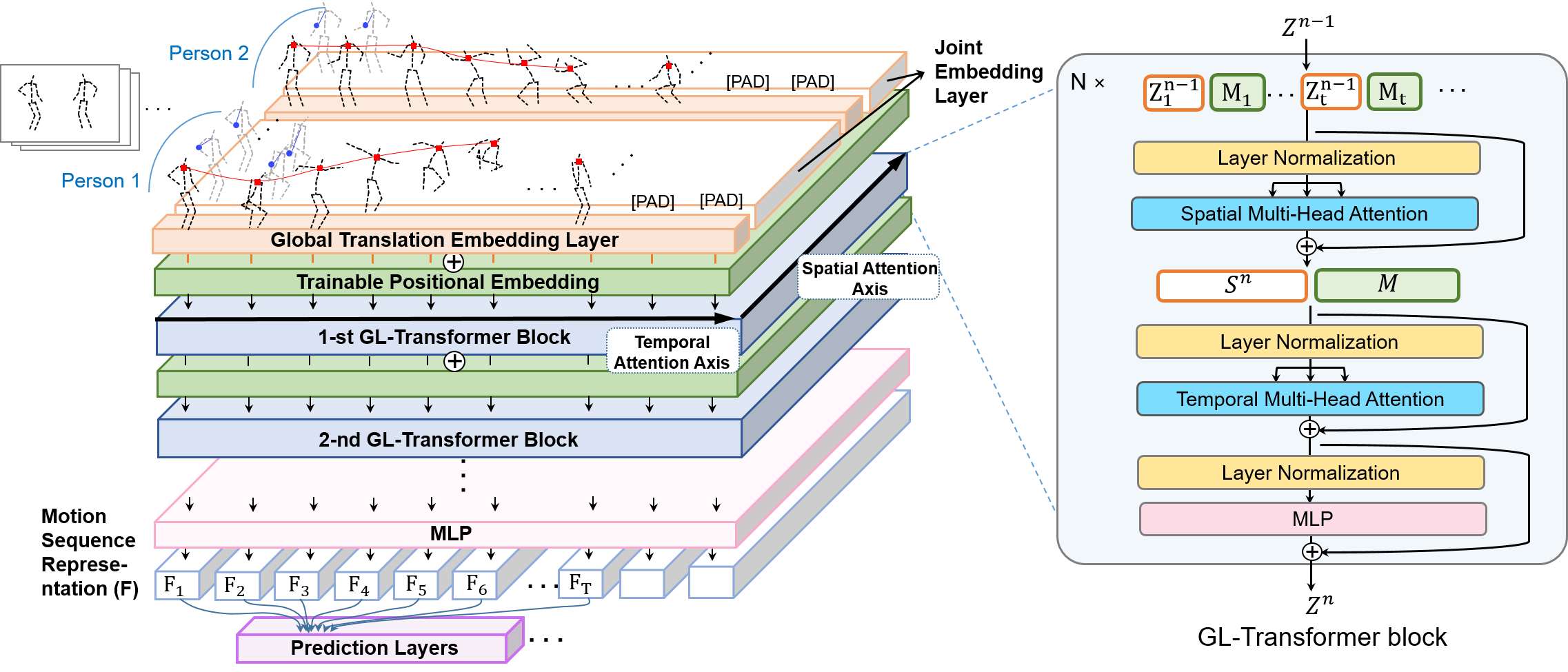}
    \vspace{-0.2cm}
    \caption{Model architecture. The input motion sequence is disentangled into global translational motions (red dots) and local motions (blue dots). The proposed model comprises $N$ stacked GL-Transformer blocks. 
    Global and local attention mechanism is implemented in both spatial-MHA and temporal-MHA modules in each block.
    }
    \label{fig:architecture}
\end{figure*}

Our model comprises $N$ stacked GL-Transformer blocks, as illustrated in Fig.~\ref{fig:architecture}, and each block contains spatial multi-head attention (spatial-MHA) and temporal multi-head attention (temporal-MHA) modules sequentially, as illustrated by the blue boxes on the right side in Fig.~\ref{fig:architecture}.
\subsubsection{Input Motion Sequences.}
As illustrated at the top of the left figure in Fig.~\ref{fig:architecture}, the input human motion sequence is expressed by two types of information: global translational motion (red dots) of the body and local motions of the body joints (blue dots).
The global translational motion represents the trajectory of the center joint of the body, and the local motions represent the relative motions of the body joints from the center joint.
The center joint is defined in each dataset, for example, NTU datasets~\cite{shahroudy2016ntu,liu2019ntu} define the spine joint as the center joint.
The original 3D skeleton motion sequence is expressed by tensor $X =[X_1, X_2, \cdots, X_T]^T$, where $X_t$ is a matrix representing the skeleton pose at the $t$-th frame.
The pose matrix $X_t$ is defined by 
$X_t = [q_t^1, q_t^2, \cdots, q_t^K]^T$, where $q_t^k \in \mathbb{R}^{3}$ indicates the 3-dimensional vector for the $k$-th joint coordinate.
The relative position of the $k$-th joint is $r_t^k = q_t^k - q_t^{c}$, where $q_t^{c}$ denotes the coordinate of the center joint.
Using the relative joint positions, the $t$-th frame of local motion is expressed by a matrix $R_t = [r_t^1, \cdots, r_t^{K}]^T$, in which we remove $r_t^{c}=(0,0,0)$ and re-index it to $K-1$ dimensional matrix as $R_t = [r_t^1, \cdots, r_t^{K-1}]^T$.
The $t$-th frame of the global translational motion is calculated using the vector $g_t = q_t^{c} - q_0^{c}$.
As in Fig.~\ref{fig:architecture}, $g_t$ and $r_t^k$ are projected into $D$ dimensional embedding vectors as
\begin{align}
    \bar{g}_t = W_g g_t + b_g, \quad \bar{r}_t^k = W_r r_t^k + b_r,  \quad k=1, \cdots, K-1, 
\end{align}
where $W_g, W_r\in\mathbb{R}^{(D \times 3)}$ and $b_g, b_r\in\mathbb{R}^{(D \times 1)}$ denote trainable weights and biases of the global translation and joint embedding layers respectively.
In the case of an action dataset containing the interaction between two or more persons, vector $g_t$ and matrix $R_t$ are expressed by $g_{t,p}$ and $R_{t,p}$ respectively, where $p$ denotes an index of the person. 
Similarly, the embedding vectors are expressed as $\bar{g}_{t,p}$, and $\bar{r}_{t,p}^k$.
In the following, we describe our method which considers the interaction among multiple persons in the sequence.

\subsubsection{Trainable and Tight Positional Embedding.}
By extending the concept of the positional embedding matrix~\cite{vaswani2017attention} containing the order information of a sequence, we introduce a trainable spatial-temporal positional embedding tensor $M\in\mathbb{R}^{T \times PK \times D}$ to learn the order information of both the temporal frames and spatial joints from the training data. Note that $PK$ is the dimension for the joint indices of $P$ persons, and $D$ is the dimension of embedding vectors, same as $D$ in $\bar{g}_{t,p}$ and $\bar{r}_{t,p}^k$.
Joint order information plays a more important role in skeleton motion sequences than in the case of sentences or images, in that
individual joint positions are not meaningful until we know which part of the body the joint belongs to.
Furthermore, frame order also plays an important role in detecting the action.
To this end, we propose a tight positional embedding method to use order information explicitly in every GL-Transformer block.
Previous transformer-based models~\cite{devlin2018bert,dosovitskiy2020image,zheng20213d} apply positional embedding once before the first transformer block.
In contrast, we apply it to the input tensors of every block, as illustrated in Fig.~\ref{fig:architecture}.
In each GL-Transformer Block, the positional embedding is explicitly applied in both the spatial-MHA and temporal-MHA modules, as illustrated in the right figure of Fig.~\ref{fig:architecture}.

\subsubsection{Global and Local Attention (GLA) Mechanism.}
We aim to construct a global and local attention (GLA) mechanism to extract global semantic information along with capturing the local relationships between the joints within the skeleton motion sequence.
GLA is implemented in both the spatial-MHA and temporal-MHA modules.
The spatial-MHA module learns spatial dependency within one frame.
In the module, global(body)-from/to-local(joint) dependencies are learned by an attention operation between the features corresponding to $g_{t,p}$ and $R_{t,p}$ of each person.
Likewise, person-to-person dependencies are learned by the attention among the features of multiple persons: $\{g_{t,p}, R_{t,p}| p=1, \cdots, P \}$, where $P$ is the number of persons. 
The temporal-MHA module learns the temporal dependencies across the sequence using pose features aggregated by the spatial-MHA. 
The temporal-MHA module learns whole-body motion information from distant frames as well as local joint dynamics from the adjacent frames.

Input pose features of the spatial-MHA module at the $t$-th frame in the $n$-th block is denoted by $Z_t^n\in\mathbb{R}^{PK\times D}$.
For the first block, embeddings of multiple people are concatenated along the spatial attention axis (see Fig.~\ref{fig:architecture}) as
\begin{align}
    Z_{t}^0 &= ||_{p=1}^P Z_{t,p}^0,  \\ 
    Z_{t,p}^0 &= [\bar{g}_{t,p}, \bar{r}_{t,p}^1, \cdots \bar{r}_{t,p}^{K-1}]^T, \quad t=1, \cdots, T,
\end{align}
where $||$ indicates the concatenation operation.
The spatial-MHA in $n( \ge 2)$-th block receives the output ($Z_t^{n-1}$) of the previous block.
The spatial-MHA module updates the pose features as
\begin{align}
S_t^{n} = \operatorname{spatial-MHA}(LN(Z_t^{n-1}+M_t)) + (Z_t^{n-1}+M_t),    
\end{align}
where $M_t\in\mathbb{R}^{PK\times D}$ is $t$-th slice of the positional embedding tensor $M$.
$LN(\cdot)$ denotes the layer normalization operator~\cite{ba2016layer}.
For the spatial-MHA$(\cdot)$, we borrow the multi-head self-attention (MHA) mechanism from~\cite{vaswani2017attention}, which is described below. 
For simplicity, we denote $LN(Z_t^{n-1}+M_t)$ as $\hat{Z}_t^{n-1}$. First, $\hat{Z}_t^{n-1}$ is projected to $query \, Q$, $key \, K$, $value \, V$ matrices as
\begin{align}
    Q = \hat{Z}_t^{n-1}W^Q, \quad K&=\hat{Z}_t^{n-1}W^K, \quad V=\hat{Z}_t^{n-1}W^V,
\end{align}
where $W^Q, W^K, W^V \in \mathbb{R}^{D \times d}$ are weight matrices for the projection and $d$ indicates the projection dimension.
The attention mechanism is expressed as
\begin{align}
    \operatorname{Attention}(Q,K,V) &= softmax(QK^T/\sqrt{d})V.
    \label{attention}
\end{align}
Note that $QK^T$ refers to the dot-product similarity of each projected joint vector in $query\, Q$ to $key\, K$. 
High attention weight is given for high similarity.
In the MHA, 
the $i$-th head performs the attention mechanism in Eq.(\ref{attention}) with different weight matrices $W_i^Q, W_i^K, W_i^V$ from those of other heads as
\begin{align}
     H_i = \operatorname{Attention}(\hat{Z}_t^{n-1}W_i^Q, &\hat{Z}_t^{n-1}W_i^K,\hat{Z}_t^{n-1}W_i^V), \quad i = 1, \cdots, h.
    \end{align}
The concatenation of $\{H_i\}$ is projected to an aggregated pose features as
\begin{align}
    \operatorname{spatial-MHA}(\hat{Z}_t^{n}) &= (||_{i=1}^h H_i)W_H, 
    \label{spatialMHA}
\end{align}
where $W_H \in \mathbb{R}^{dh \times dh}$ is a projection matrix. 

To perform temporal-MHA in the $n$-th block, we vectorize the pose feature of the $t$-th frame $S_t^n \in R^{PK \times D}$ into $s_t^n \in R^{PK\cdot D}$. Then, the vectorized pose features are stacked to form a pose feature sequence matrix $S^n=[s_1^{n}, s_2^{n}, \dots, s_T^{n}]^T$ $\in\mathbb{R}^{T\times (PK \cdot D)}$. 
In the temporal-MHA module, the same MHA mechanism in Eq.(\ref{spatialMHA}) is used, but different weight matrices are applied.
Then, the output pose sequence feature of the $n$-th GL-Transformer ($Z^n$) is obtained through MLP($\cdot$), that is, 
\begin{align}
    \bar{Z}^{n} &= \operatorname{temporal-MHA}(LN(S^n+\bar{M})) + (S^n+\bar{M}), \\ 
    Z^{n} &= MLP(LN(\bar{Z}^{n})) + \bar{Z}^{n}, 
\end{align}
where $\bar{M} \in\mathbb{R}^{T \times (PK \cdot D)}$ is a matrix in which the dimension of the positional embedding tensor $M \in\mathbb{R}^{T \times PK \times D}$ is changed.
In the $N$-th GL-Transformer block, the final motion sequence representation $F$ for the input motion sequence $X$ is obtained by passing $Z^{N}$ through a 2-layer MLP as
\begin{align}
    F=\operatorname{GL-Transformer}(X) = MLP(Z^{N}).
    \label{final_representation}
\end{align}

\subsubsection{Masked Attention for Natural-Speed Motion Sequence.}
Most of the existing action recognition methods~\cite{cheng2021hierarchical,su2020predict,li20213d,yang2021skeleton}
employ a fixed length of motion sequences, which overlooks the importance of the speed of the motion.
To handle natural-speed motion sequences, we utilize an attention mask~\cite{vaswani2017attention}, so that our model can learn the natural joint dynamics across frames and capture speed characteristics from diverse actions.
To this end, we define the maximum sequence length as $T_{max}$.
If the length of the original sequence $X_{ori}$ is shorter than $T_{max}$, the rest of the frames are filled with padding dummy tokens $[PAD]\in \mathbb{R}^{PK \times 3}$, which yields $X =[X_{ori}^T, [PAD], ..., [PAD]]^T \in \mathbb{R}^{T_{max} \times PK \times 3}$.
Elements of $[PAD]$ are set to arbitrary numbers because the loss corresponding to the $[PAD]$ token is excluded.
To exclude attention from the dummy values, we mask (setting to $-\infty$) columns corresponding to the $[PAD]$ tokens in the $QK^T$ matrix.

\subsection{Multi-interval Pose Displacement Prediction (MPDP) Strategy}
\label{sec:pretraining_strategy}
We design a novel pretraining strategy, multi-interval pose displacement prediction (MPDP), which estimates the whole-body and joint motions at various time intervals at different scales.
H-transformer~\cite{cheng2021hierarchical} introduces a pretraining strategy that estimates the direction of the instantaneous joint velocity.
The instantaneous velocity of the joint in a specific frame can be easily obtained from the adjacent frame so that the model is guided to learn local attention rather than long-range global attention.
To overcome this limitation, we propose an MPDP strategy to effectively learn global attention as well as local attention.

\begin{figure*}[t]
\centering
    \includegraphics[width=12cm]{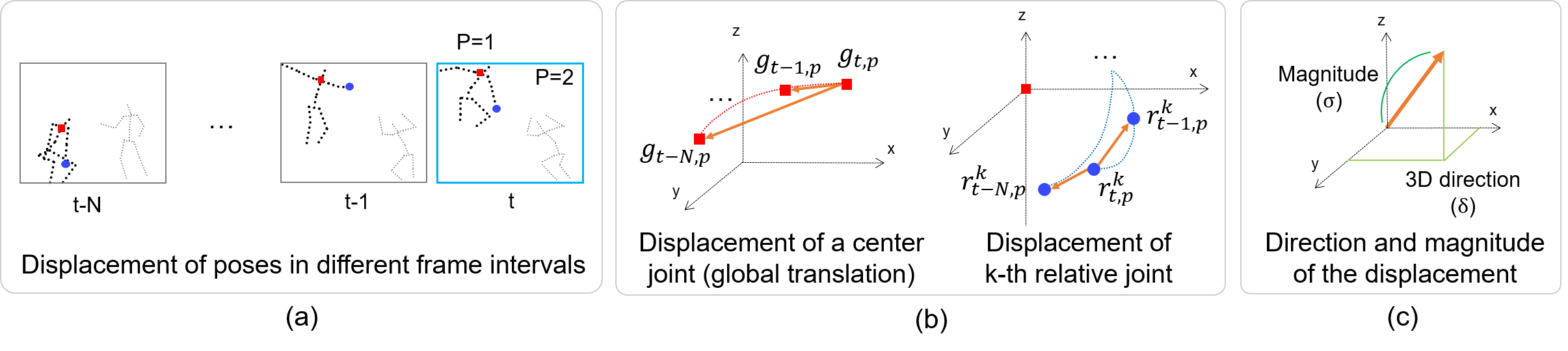}
    \vspace{-0.3cm}
    \caption{Description of multi-interval pose displacement}
    \label{fig:pretraining_task}
\end{figure*}

As illustrated in Fig.~\ref{fig:pretraining_task} (a), we first select multiple frame intervals $t-N, \cdots, t-n, \cdots, t$.
GL-Transformer is trained to predict the magnitude and direction of the pose displacement between the $t$-th and $(t-n)$-th frame.
Local motion (relative joint displacement) is predicted with the help of global motion and vice versa. In addition, the motion of other people is considered when predicting one's motion.
The displacements are represented by the orange arrows in Fig.~\ref{fig:pretraining_task} (b) and (c).
We design the pose displacement prediction as a classification task using {\it softmaxed} linear classifiers.
The model is trained to predict both the direction and magnitude classes for each interval.
The predictions of the $t$-th frame for the interval $n$ are expressed as
\begin{align}
    \hat{\Delta}_{t,n} = softmax(W_n^{\delta}F_t+b_n^{\delta}), \quad \hat{\Sigma}_{t,n} = softmax(W_n^{\sigma}F_t+b_n^{\sigma}),
\end{align}
where $F_t$ denotes $t$-th slice of the motion sequence representation $F$, as shown in the left side of Fig.~\ref{fig:architecture}.
$\hat{\Delta}_{t,n} =||_{p=1}^P \hat{\Delta}_{t,p,n}$ where 
$\hat{\Delta}_{t,p,n}=
[\hat{\delta}_{t,p,n}^g, \hat{\delta}_{t,p,n}^1, \cdots, \hat{\delta}_{t,p,n}^{K-1}]^T$,
and $\hat{\delta}_{t,p,n}^g,$ $\hat{\delta}_{t,p,n}^k\in\mathbb{R}^{C_{\delta}}$ denotes the predicted direction class vector of global translation and $k$-th joints, respectively. 
$C_\delta$ is the number of direction classes.  
$\hat{\Sigma}_{t,n}=||_{p=1}^P \hat{\Sigma}_{t,p,n}$ where 
$\hat{\Sigma}_{t,p,n}=
[\hat{\sigma}_{t,p,n}^g, \hat{\sigma}_{t,p,n}^1, \cdots, \hat{\sigma}_{t,p,n}^{K-1}]^T$,
and $\hat{\sigma}_{t,p,n}^g, \hat{\sigma}_{t,p,n}^k\in\mathbb{R}^{C_{\sigma}}$ denotes the predicted magnitude class vector of the global translation and $k$-th joints, respectively. $C_\sigma$ denotes the number of magnitude classes.
$W_n^{\delta},W_n^{\sigma},b_n^{\delta}$, and $b_n^{\sigma}$ are the trainable weights and biases of the linear classifiers for interval $n$.
To train the model parameters, we define the ground truth classes of direction $\delta$ and magnitude $\sigma$ at the $t$-th frame for the $p$-th person and interval $n$ as
\begin{align}
    \delta_{t,p,n}^g &= class(\angle(g_{t,p}-g_{(t-n),p})), \quad \delta_{t,p,n}^k = class(\angle(r_{t,p}^k-r_{(t-n),p}^k)), \\
    \sigma_{t,p,n}^g &= class(\lVert g_{t,p}-g_{(t-n),p} \rVert), \quad \sigma_{t,p,n}^k = class(\lVert r_{t,p}^k-r_{(t-n),p}^k \rVert),
\end{align}
where we set $g_{(t-n),p} = g_{t,p}$ and $r_{(t-n),p}^k=r_{t,p}^k$ at $t \leq n$ because we do not have the information of the $(t-n)$-th frame in this case.
$class(\cdot)$ denotes the class label vector of $\cdot$, where the magnitude is quantized into one of the $C_{\sigma}$ classes, and the direction is designated as one of the $C_{\delta}=27$ classes, in which each of the xyz direction has three classes: +,$-$, and no movement.
The classification loss is calculated for all intervals and frames except the [PAD] tokens.
The total loss is defined as follows:
\begin{align}
    L_{total} &=  \sum_{t=1}^{T} \sum_{p=1}^{P} \sum_{n}^{N}\Big(\lambda_{\delta} L_{\delta}(t,p,n) + \lambda_{\sigma} L_\sigma(t,p,n) \Big), 
\end{align}
where direction loss $L_{\delta}(t,p,n)$ and magnitude loss $L_{\sigma}(t,p,n)$ are the weighted sum of cross entropy loss to train each component of $\hat{\Delta}_{t,p,n}$ and $\hat{\Sigma}_{t,p,n}$, whereas 
$\lambda_\delta$ and $\lambda_\sigma$ denote the weighting factors of $L_\delta$ and $L_\sigma$, respectively.
\section{Experiments}
\subsection{Datasets \& Evaluation Protocol}
\textbf{NTU-RGB+D.}
NTU-RGB+D 60 (NTU-60)~\cite{shahroudy2016ntu} is a large-scale dataset containing 56,880 3D skeleton motion sequences performed by up to two actors and categorized into 60 action classes.
Each person has 25 joints.
We follow two standard evaluation criteria: cross-subject (\textbf{xsub}) and cross-view (\textbf{xview}).
In \textbf{xsub}, the training and test set are collected by different subjects. \textbf{xview} splits the training and testing set according to the camera view.
NTU-RGB+D 120 (NTU-120)~\cite{liu2019ntu} is an extension of NTU- 60 which contains 113,945 sequences for 120 action classes. 
The new evaluation criterion cross-setup (\textbf{xset}) is added for NTU-120, whose training and testing sets are split by the camera setup IDs.
\\
\textbf{North-Western UCLA.}
North-Western UCLA (NW-UCLA)~\cite{wang2014cross} contains 1,494 motion sequences captured  by 10 subjects.
Each sequence is performed by one actor and each person has 20 joints. The actions are categorized into 10 action classes. 
Following the standard evaluation protocol, the training set comprises samples from camera views 1 and 2, and the remaining samples from view 3 are arranged in the testing set.
\\
\textbf{Evaluation Protocol.}
We adopt linear evaluation protocol~\cite{zheng2018unsupervised,lin2020ms2l,xu2021prototypical,rao2021augmented,kundu2019unsupervised,nie2020unsupervised,cheng2021hierarchical} which is the standard for the evaluation of unsupervised learning tasks.
Following the protocol, the weight parameters of the pretrained model are fixed, and only the attached single linear classifier is trained with the training data.
In addition, we evaluate the proposed model in semi-supervised settings~\cite{su2021self,si2020adversarial,li20213d,yang2021skeleton}.
The pretrained model is fine-tuned with 5\% and 10\% of the training data, and then the action recognition accuracy is evaluated.

\subsection{Implementation Details}
We set $T_{max}=300$ for the NTU dataset and $T_{max}=50$ for the NW-UCLA dataset. 
The sequence is augmented by applying a shear~\cite{rao2021augmented} and interpolation. 
For interpolation, the sequence is interpolated into a random length within $\pm$ 10\% of the original sequence length.
Since the NTU dataset includes two persons, we set it to $P=2$. 
Four transformer blocks are utilized,  the hidden dimension $D=6$ for each joint, and eight heads ($h=8$) are used for self-attention.
The H-transformer~\cite{cheng2021hierarchical} uses four transformer blocks with $D=256$ for each of the five body parts.
We set $\lambda_\delta, \lambda_\sigma=1$.
In the unsupervised pretraining phase, we utilize the AdamW~\cite{loshchilov2017decoupled} optimizer with an initial learning rate of $5e^{-4}$ and decay it by multiplying by 0.99 every epoch. The model is trained for 120 epochs for the NTU and 300 epochs for the NW-UCLA with a batch size of 128.
In the linear evaluation protocol, we utilize Adam~\cite{kingma2014adam} optimizer with a learning rate of $3e^{-3}$. 
The linear layer is trained for 120 and 300 epochs for NTU and NW-UCLA, respectively, with  a batch size of 1024.

\subsection{Ablation Study}
We conduct ablation studies using the NTU-60 dataset to demonstrate the effectiveness of the main components of our method.
The final performance of GL-Transformer substantially exceeds that of the  H-transformer~\cite{cheng2021hierarchical} in the linear evaluation protocol, exceeding 7.0\% for \textbf{xsub} and 11.0\% for \textbf{xview}.
The effectiveness of each component is explained as follows:
\\
\textbf{Effectiveness of GLA and MPDP.}
In Table~\ref{table:ablation-global}, Experiment \textbf{(1)} exploits the original pose sequence $X$, Experiment \textbf{(2)} utilizes local motion $R_t (t=1,\cdots,T)$, and Experiment \textbf{(3)} utilizes both global translational motion $g_t (t=1,\cdots,T)$ and local motion  $R_t (t=1,\cdots,T)$.
Regarding \textbf{(1)}, because global and local motions are mixed in $X$, it is difficult to model both global and local motions.
The result of \textbf{(2)} is higher than that of \textbf{(1)} when the model learns local dynamics between the joints from local motions.
The result of \textbf{(3)} is further improved demonstrating that GLA plays an important role in extracting the representation of the entire motion sequence effectively. 
\begin{table}[t]
\caption{Ablation study for verifying the effectiveness of GLA and MPDP in the NTU-60 dataset with the linear evaluation protocol}
\label{table:ablation-global}
\centering
\resizebox{0.95\textwidth}{!}{
\begin{tabular}{ccccc|DD} 
\toprule
\multicolumn{1}{c}{\multirow{2}{*}{\begin{tabular}[c]{@{}c@{}}\\ \end{tabular}}} &
\multicolumn{1}{c}{\multirow{2}{*}{\begin{tabular}[c]{@{}c@{}}Displacement\\ direction\end{tabular}}} & \multicolumn{1}{c}{\multirow{2}{*}{\begin{tabular}[c]{@{}c@{}}Disentangle\\ global translation\end{tabular}}} &
\multirow{2}{*}{\begin{tabular}[c]{@{}c@{}}Displacement\\ magnitude\end{tabular}} & 
\multirow{2}{*}{\begin{tabular}[c]{@{}c@{}}Frame\\ interval\end{tabular}} & 
\multicolumn{2}{c}{Accuracy(\%)}\\
 & & & & & \multicolumn{1}{c}{xsub} & \multicolumn{1}{c}{xview} \\ \midrule

H-transformer~\cite{cheng2021hierarchical}& \checkmark  &           &           & \{1\}           &   69.3    &  72.8 \\ \midrule
Experiment \textbf{(1)}&\checkmark  &            &              & \{1\}           &   71.1            &   73.5  \\
Experiment \textbf{(2)}&\checkmark  &\checkmark(only local motion) &              & \{1\}           &   74.2            &   81.9  \\
Experiment \textbf{(3)}&\checkmark  &\checkmark  &              & \{1\}           &   75.4            &   82.8  \\
Experiment \textbf{(4)}&\checkmark  &\checkmark  &\checkmark    & \{1\}           &   75.7            &   82.9  \\
Experiment \textbf{(5)}&\checkmark  &\checkmark  &\checkmark    & \{1, 5\}        &   75.9            &   83.3  \\
Experiment \textbf{(6)}&\checkmark  &\checkmark  &\checkmark    & \{1, 5, 10\}    &  \textbf{76.3}    &   \textbf{83.8}  \\
Experiment \textbf{(7)}&\checkmark  &\checkmark  &\checkmark    & \{1, 5, 10, 15\}&   75.7            &   83.4  \\

\bottomrule
\end{tabular} 
}
\end{table}
\begin{table}[t]
    \caption{Ablation study for verifying the effectiveness of person-to-person attention in NTU-120 \textbf{xsub} (left) and trainable and tight positional embedding (right) in NTU-60 with the linear evaluation protocol }
    \vspace{-0.2cm}
    \begin{minipage}{0.55\linewidth}
    \label{table:ablation-p2p-posemb}
    \centering
        \resizebox{0.95\textwidth}{!}{
        \begin{tabular}{c|cc|D}
        \toprule
        p2p & \multicolumn{3}{c}{Accuracy(\%)}\\ 
        attention      & one person cat.   & two people cat.&  total  \\\midrule
        w/o       & 63.0  &   71.6    &   64.9    \\
        w/$~~$    & {\bf 63.7}  & {\bf 73.5}  & {\bf 66.0} \\
        \bottomrule
        \end{tabular} 
        }
    \end{minipage}%
    \begin{minipage}{0.45\linewidth}
    \centering
        \resizebox{0.85\textwidth}{!}{
        \begin{tabular}{l|DD}
        \toprule
        \multirow{2}{*}{Type} & \multicolumn{2}{c}{Accuracy(\%)}\\ 
                              & xsub      & xview              \\\midrule
        Fixed (sinusoidal)              & 75.5    &   83.3    \\
        Trainable                       & 76.0    &   83.6    \\
        Trainable tight                 & \textbf{76.3}    &  \textbf{83.8}   \\
        \bottomrule
        \end{tabular} 
        }
    \end{minipage} 
\end{table}

In Table~\ref{table:ablation-global}, Experiment \textbf{(3)} does not adopt the displacement magnitude prediction loss, that is $\lambda_\sigma=0$. 
For Experiment \textbf{(4)}, $\lambda_\sigma=1$, and predicting both directions and magnitudes exhibits a higher performance.
Experiments \textbf{(4)} to \textbf{(7)} are performed by altering the frame intervals that are utilized in MPDP.
The performance gradually increases from the interval $n=\{1\}$ to $n=\{1,5,10\}$, demonstrating that long-range global attention is effective in aggregating the context of the entire motion sequence.
The accuracy corresponding to the interval $n=\{1,5,10,15\}$ is lower than that of $n=\{1,5,10\}$.
This implies the maximum interval relies on the inter-frame dependency of the given sequence.
\\
\textbf{Effectiveness of Person-to-person Attention.}
To verify the effect of person-to-person (p2p) attention, we report the model performance trained with and without p2p attention in Table~\ref{table:ablation-p2p-posemb}. NTU-120 has 120 action categories, 26 among them are two-person interactions and the rest are one-person actions.
The p2p attention improves the performance of both groups, especially the performance increases more in the group of two people categories.
\\
\textbf{Effectiveness of Trainable Tight Positional Embedding.}
For positional embedding, the performance increases when a trainable embedding is employed instead of a fixed sinusoidal embedding, as presented in the right table of Table~\ref{table:ablation-p2p-posemb}.
The use of tight embedding further increases the performance.
We also verify that frames close to each other are trained to have similar positional embeddings.
The corresponding figures are added in the supplementary material.
In addition, the experiment demonstrating the effectiveness of natural-speed input is added to the supplementary material.

\subsection{Analysis of Learned Attention}
\label{sec:analysis_of_attention}
\begin{figure*}[t] 
\centering
    \includegraphics[width=12.0cm]{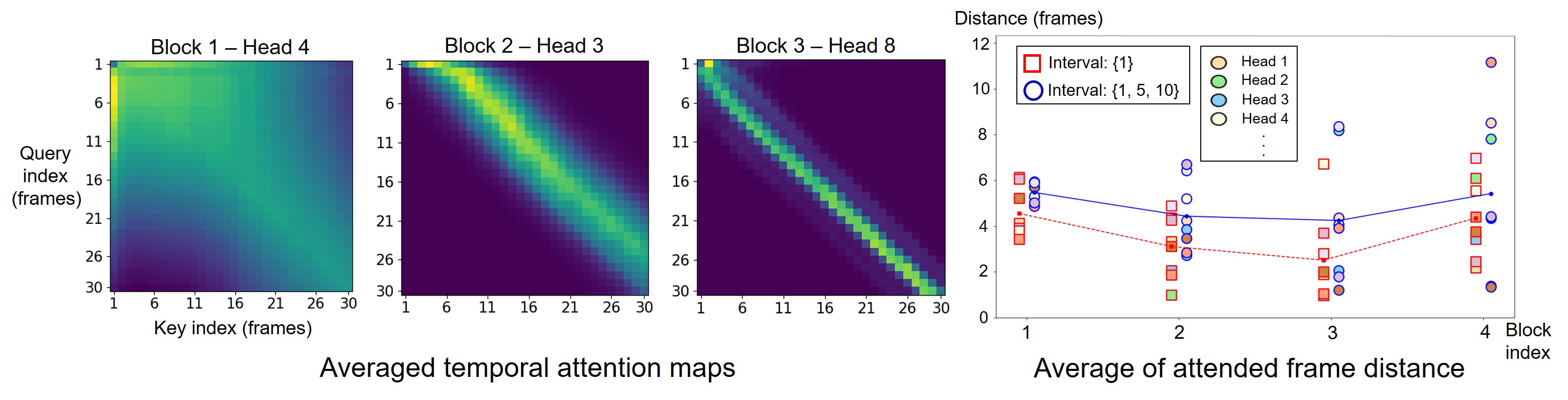}
    \vspace{-0.4cm}
    \caption{Examples of learned temporal attention maps averaged over 300 evaluation sequences (left) and average of attended frame distance (right). Yellow color indicates a large value in the left figure. Blue (interval \{1,5,10\}) and red (interval \{1\}) lines indicate the average values over heads in each block}
    \label{fig:temporal_attention_maps}
\end{figure*}

\begin{figure*}[t]
\centering
    \includegraphics[width=12.0cm]{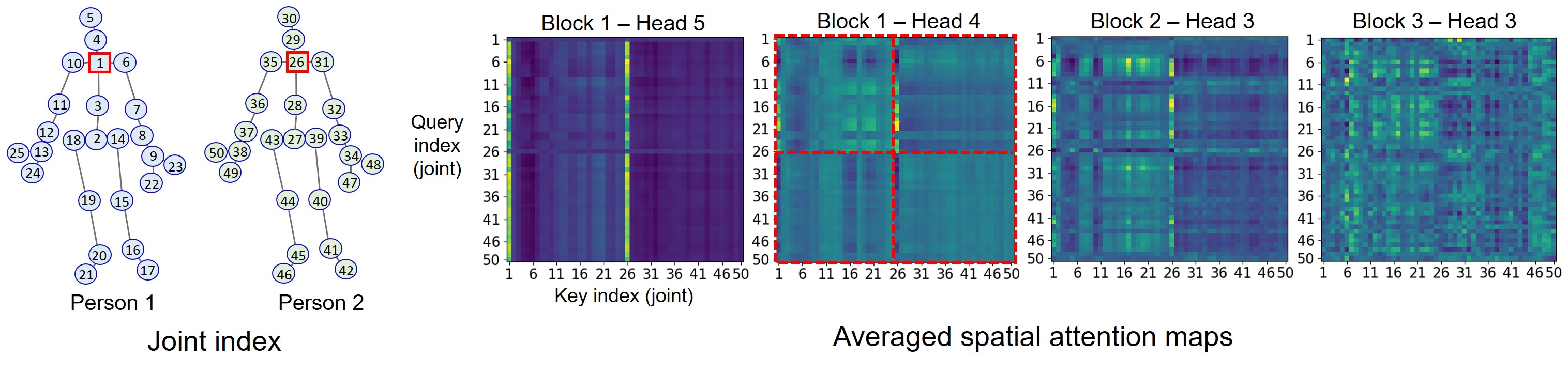}
    \vspace{-0.4cm}
    \caption{Examples of learned spatial attention maps averaged over 300 evaluation sequences. Yellow color indicates a large value}
    \label{fig:spatial_attention_maps}
\end{figure*}

We analyze the attention map, $softmax(QK^T/\sqrt{d})$ in Eq.(\ref{attention}), of each pretrained GL-Transformer block.
The spatial and temporal attention maps are extracted from the spatial-MHA and temporal-MHA modules, respectively. 
The attention maps are averaged over 300 motion sequences from the evaluation data.
Each head of each transformer block indicates various types of attention maps, and representative samples are shown in Fig.~\ref{fig:temporal_attention_maps} and Fig.~\ref{fig:spatial_attention_maps}.
In Fig.~\ref{fig:temporal_attention_maps}, we indicate the averaged temporal attention map for the first 30 frames, because the length of test sequences varies from each other.
The vertical and horizontal axes represent the \textit{query} and \textit{key} indices, respectively, and the color of each pixel indicates the degree to which the \textit{query} attends to the \textit{key}.
Each head attends a different temporal range, for example, approximately neighboring 10 frames and 5 frames are highlighted in the attention maps of Block2-Head3 and Block3-Head8, respectively, whereas a wide range is highlighted in the attention map of Block1-Head4.
The figure on the right in Fig.~\ref{fig:temporal_attention_maps} illustrates the average attended frame distances~\cite{dosovitskiy2020image} of each head.
The average of the attended frame distances~\cite{dosovitskiy2020image} is calculated as a weighted sum of the frame distances, where attention is regarded as the weight. Red squares indicate each head when using frame interval $\{1\}$, and blue circles indicate each head when using intervals $\{1,5,10\}$.
In each block, more heads attend to distant frames when the model is pretrained with intervals $\{1,5,10\}$ as compared to when the model is pretrained with interval $\{1\}$.

An example of the spatial attention map is illustrated in Fig.~\ref{fig:spatial_attention_maps}.
The $1$-st and $26$-th indices are utilized for the global translations corresponding to $g_{t,1}$ and $g_{t,2}$, respectively, which are represented as red squares in the left figure.
In some heads, the $1$-st and $26$-th indices appear to be attended differently from other joints.
For example, in Block1-Head5, \textit{queries} of all joints pay attention to the $1$-st and $26$-th \textit{keys} of more than \textit{keys} of other joints.
In Block1-Head4, correlations between the joints corresponding to each person are observed as 4 divisions in the attention map, as indicated in red dotted lines.
Overall, the proposed model learns the global relationships at shallow blocks (i.e. Block1) and learns fine-grained relationships at deeper blocks (i.e. Block2 and Block3).

\subsection{Comparison with State-of-the-art Methods}
We compared our approach with the state-of-the-art (SOTA) methods for unsupervised action recognition: methods using RNN-based encoder-decoder models~\cite{zheng2018unsupervised,su2020predict,lin2020ms2l,xu2021prototypical,kundu2019unsupervised,nie2020unsupervised}, a method using GRU-based encoder-decoder model~\cite{yang2021skeleton}, methods using contrastive learning scheme~\cite{rao2021augmented,li20213d}, and a transformer-based method~\cite{cheng2021hierarchical}.
We use a linear evaluation protocol to measure the action recognition accuracy.
The performance of our method substantially exceeds that of the H-transformer~\cite{cheng2021hierarchical} which focuses only on the local relationship between body parts and between frames.
As presented in Table~\ref{table:sota-ntu60}, the performance of GL-Transformer exceeds that of the H-transformer by $7.0\%$ in \textbf{xsub} and $11.0\%$ in \textbf{xview} in the NTU-60 dataset.
Furthermore, our method outperforms all previous methods by a notable margin.

\begin{table}[t]
\caption{Action recognition results with linear evaluation protocol in NTU-60 dataset}
\label{table:sota-ntu60}
\centering
\resizebox{0.88\textwidth}{!}{
\begin{tabular}{L|B|DD}
\toprule
\multirow{2}{*}{Method} & \multirow{2}{*}{Network}  & \multicolumn{2}{c}{Accuracy(\%)}\\ 
                        &                           & xsub      & xview              \\\midrule
LongT GAN (2018)~\cite{zheng2018unsupervised}       & GRU (encoder-decoder)         &   39.1    &   48.1    \\
P\&C (2020)~\cite{su2020predict}                    & GRU (encoder-decoder)         &   50.7    &   76.3    \\
MS$^2$L (2020)~\cite{lin2020ms2l}                   & GRU (encoder-decoder)         &  52.6    &   -       \\
PCRP (2021)~\cite{xu2021prototypical}               & GRU (encoder-decoder)         &   54.9    &   63.4    \\
AS-CAL (2021)~\cite{rao2021augmented}               & LSTM (contrastive learning)   &   58.5    &   64.6    \\
CRRL (2021)~\cite{wang2021contrast}                 & LSTM (contrastive learning)   &   67.6    &   73.8    \\
EnGAN-PoseRNN (2019)~\cite{kundu2019unsupervised}   & RNN (encoder-decoder)         &   68.6    &   77.8    \\
SeBiReNet (2020)~\cite{nie2020unsupervised}         & GRU (encoder-decoder)         &   -       &   79.7    \\
`TS' Colorization (2021)~\cite{yang2021skeleton}    & GCN (encoder-decoder)         &   71.6    &   79.9    \\
CrosSCLR-joint (2021)~\cite{li20213d}               & GCN (contrastive learning)    &   72.9    &   79.9    \\
CrosSCLR-bone (2021)~\cite{li20213d}               & GCN (contrastive learning)     &   75.2    &   78.8    \\
\midrule
H-transformer (2021)~\cite{cheng2021hierarchical}   &Transformer                    &   69.3    &   72.8    \\
\textbf{GL-Transformer}                                  &Transformer                    &   \textbf{76.3}    &   \textbf{83.8} \\
\bottomrule
\end{tabular} 
}
\end{table}


On the NTU-120 dataset, GL-Transformer outperforms the SOTA methods with a significant margin, as presented in the left table of  Table~\ref{table:sota-ntu120-ucla}.
It is verified that the proposed method operates robustly on datasets that include more detailed actions.  
On the NW-UCLA dataset, GL-Transformer achieved the highest performance among the previous methods, demonstrating that the proposed model is effective even with a small amount of training data, as presented in the right table of Table~\ref{table:sota-ntu120-ucla}.
In addition, we compare the results from the semi-supervised setting on the NTU-60 and NW-UCLA datasets in Table~\ref{table:sota-finetune}.
The results of the SOTA semi-supervised action recognition methods~\cite{si2020adversarial,su2021self} are also compared in conjunction with the unsupervised methods aforementioned.
GL-Transformer exceeds SOTA performance in both evaluations using 5\% and 10\% of the training data.
\begin{table}[t]
\caption{Action recognition results with linear evaluation protocol in the NTU-120 dataset (left) and NW-UCLA dataset (right)}
\vspace{-0.2cm}
    \begin{minipage}{0.5\linewidth}
    \label{table:sota-ntu120-ucla}
    \centering
        \resizebox{0.9\textwidth}{!}{
        \begin{tabular}{l|DD}
        \toprule
        \multirow{2}{*}{Method} & \multicolumn{2}{c}{Accuracy(\%)}\\ 
                              & xsub      & xset              \\\midrule
        P\&C (2020)~\cite{su2020predict}                &   41.7    &   42.7    \\
        PCRP (2021)~\cite{xu2021prototypical}           &   43.0    &   44.6    \\
        AS-CAL (2021)~\cite{rao2021augmented}           &   48.6    &   49.2    \\
        CrosSCLR-bone (2021)~\cite{li20213d}            &   53.3    &   50.6    \\
        CRRL (2021)~\cite{wang2021contrast}             &   56.2    &   57.0    \\
        CrosSCLR-joint (2021)~\cite{li20213d}           &   58.8    &   53.3    \\
        \midrule
        \textbf{GL-Transformer}                              &   \textbf{66.0}    &   \textbf{68.7} \\
        \bottomrule
        \end{tabular} 
        }
    \end{minipage}%
    \begin{minipage}{0.5\linewidth}
    \centering
        \resizebox{0.8\textwidth}{!}{
        \begin{tabular}{l|cc}
        \toprule
        Method                  & Accuracy(\%)  \\\midrule
        LongT GAN (2018)~\cite{zheng2018unsupervised}       &   74.3 \\
        MS$^2$L (2020)~\cite{lin2020ms2l}                   &   76.8 \\
        SeBiReNet (2020)~\cite{nie2020unsupervised}         &   80.3 \\
        CRRL (2021)~\cite{wang2021contrast}                 &   83.8 \\
        P\&C (2020)~\cite{su2020predict}                    &   84.9 \\
        PCRP (2021)~\cite{xu2021prototypical}               &   86.1 \\
        `TS' Colorization (2021)~\cite{yang2021skeleton}    &   90.1 \\
        \midrule
        H-transformer (2021)~\cite{cheng2021hierarchical}   &   83.9  \\
        \textbf{GL-Transformer}                                  & \textbf{90.4}\\
        \bottomrule
        \end{tabular}
        }
    \end{minipage} 
\end{table}

\begin{table}[t]
\caption{Results with semi-supervised setting in the NTU-60 and NW-UCLA datasets}
\vspace{-0.1cm}
\label{table:sota-finetune}
\centering
\resizebox{0.9\textwidth}{!}{
\renewcommand{\tabcolsep}{1.5mm}
\begin{tabular}{l|DD|DD|DD}
\toprule
\multirow{2}{*}{Methods} & \multicolumn{2}{c|}{NTU-60 (xsub)} & \multicolumn{2}{c|}{NTU-60 (xview)} & \multicolumn{2}{c}{NW-UCLA} \\
                         & 5\%              & 10\%              & 5\%               & 10\%              & 5\%           & 10\%  \\ \midrule
MCC-ST-GCN (2021)~\cite{su2021self}                 & 42.4      & 55.6      & 44.7      & 59.9      & -     & -     \\
MCC-2s-AGCN (2021)~\cite{su2021self}                & 47.4      & 60.8      & 53.3      & 65.8      & -     & -     \\
MCC-AS-GCN (2021)~\cite{su2021self}                 & 45.5      & 59.2      & 49.2      & 63.1      & -     & -     \\
LongT GAN (2018)~\cite{zheng2018unsupervised}       & -         & 62.0      & -         & -         & -     & 59.9  \\
ASSL (2020)~\cite{si2020adversarial}                & 57.3      & 64.3      & 63.6      & 69.8      & 52.6  & -     \\
MS$^2$L (2020)~\cite{lin2020ms2l}                   & -         & 65.2      & -         & -         & -     & 60.5  \\
CrosSCLR-bone (2021)~\cite{li20213d}                & 59.4      & 67.7      & 57.0      & 67.3      & -     & -  \\
'TS' Colorization (2021)~\cite{yang2021skeleton}    & 60.1      & 66.1      & 63.9      & 73.3      & 55.9  & 71.3  \\
CrosSCLR-joint (2021)~\cite{li20213d}               & 61.3      & 67.6      & 64.4      & 73.5      & -     & -  \\
\midrule
\textbf{GL-Transformer}                                  &\textbf{64.5}      & \textbf{68.6}      & \textbf{68.5}       & \textbf{74.9}        & \textbf{58.5}    & \textbf{74.3}\\
\bottomrule
\end{tabular}
}
\end{table}

\section{Conclusions}
We introduce a novel transformer architecture and pretraining strategy suitable for motion sequences.
The proposed GL-Transformer successfully learns global and local attention, so that the model effectively captures the global context and local dynamics of the sequence. 
The performance of our model substantially exceeds those of SOTA methods in the downstream action recognition task in both unsupervised and self-supervised manners.
In future studies, our model can be extended to a model for learning various skeleton features together, such as the position and bone, to encode richer representations.
The memory usage and computation of the model are expected to be reduced by using the concept of sparse attention~\cite{child2019generating,beltagy2020longformer}, which sparsely pays attention to each other among tokens.
Furthermore, our model can be extended to a large-parameter model and pretrained with a large number of skeleton sequences extracted from unspecified web videos to be more generalized, and can be applied to various downstream tasks dealing with human actions.

\vspace{-0.2cm}
\subsubsection*{Acknowledgement.}
\footnotesize{This work was supported by IITP/MSIT [B0101-15-0266, Development of high performance visual bigdata discovery platform for large-scale realtime data analysis, 1/4; 2021-0-01343, AI graduate school program (SNU), 1/4; 2021-0-00537, Visual common sense through self-supervised learning for restoration of invisible parts in images, 1/4; 1711159681, Development of high-quality AI-AR interactive media service through deep learning-based human model generation technology, 1/4]}

\clearpage
\bibliographystyle{splncs04}
\bibliography{egbib}

\clearpage

\setcounter{section}{0}
\setcounter{figure}{0}
\setcounter{table}{0}
\renewcommand\thesection{\Alph{section}}
\renewcommand{\thefigure}{\Alph{figure}}
\renewcommand{\thetable}{\Alph{table}}


\begin{center}
\Large\textbf{Supplementary Material for \\ Global-local Motion Transformer for Unsupervised Skeleton-based Action Learning}
\end{center}

\section{Learned Positional Embedding}
\begin{figure*}[ht]
\centering
    \includegraphics[width=17.0cm, angle=90]{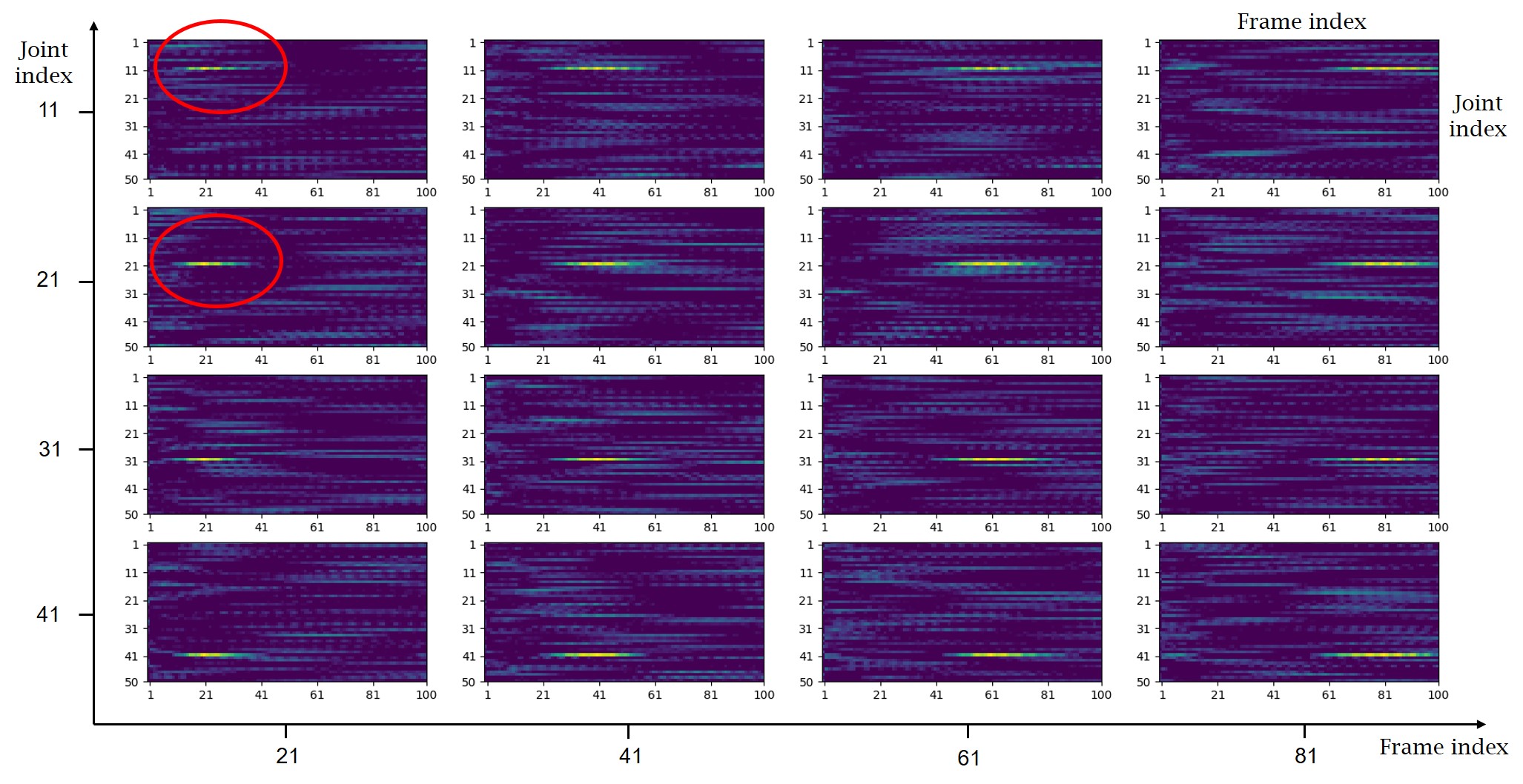}
    \caption{Similarity between learned positional embedding vectors.
    Each tiled figure shows the cosine similarity between the positional embedding vector of the indicated joint and frame and that of all other indices.
    Yellow color indicates a large cosine similarity}
    \label{fig:supple_positional_embedding}
\end{figure*}
As illustrated in Fig.~\ref{fig:supple_positional_embedding}, the positional embedding obtained through training effectively reflects the sequential information of the motion sequence.
Fig.~\ref{fig:supple_positional_embedding} presents similarity between learned positional embedding vectors $M_{t,j}\in\mathbb{R}^{D}$, where $t=1,\cdots,T$ and $j=1,\cdots, PK$.
$M_{t,j}$ is a slice of the positional embedding tensor $M\in\mathbb{R}^{T\times PK \times D}$, where $t$ and $j$ denote the frame and joint index, respectively.
Each tiled figure shows a cosine similarity between the positional embedding vector of the indicated joint and frame and that of all other indices.
As indicated as red circles, closer frames tend to have similar positional embedding vectors. 
This implies that the positional embedding is effectively trained to reflect temporal order within the motion sequence.

\section{Effectiveness of Natural-speed Input}
The performance of using the input sequence with a natural speed exceeds the performance of using the sampled 150 and 300 frames as an input, as presented in Table~\ref{table:ablation-input}.
The results verify that the proposed scheme is effective in learning the natural dynamics between joints.
Note that H-tarnsformer~\cite{cheng2021hierarchical} leverages the input sequences as sampling to 150 frames.
The result of sampling 300 frames is worse than that of sampling 150 frames, owing to the large distortion of the motion sequences.
\begin{table}[b]
\vspace{-0.2cm}
    \caption{Ablation study for verifying the effectiveness of natural-speed input sequences in the NTU-60 dataset with the linear evaluation protocol }

    \label{table:ablation-input}
    \centering
        \resizebox{0.6\textwidth}{!}{
        \begin{tabular}{l|CC}
        \toprule
        \multirow{2}{*}{Type} & \multicolumn{2}{c}{Accuracy(\%)}\\ 
                              & xsub      & xview              \\\midrule
        H-transformer~\cite{cheng2021hierarchical} &   69.3    &   72.8    \\ \midrule
        Sampling (150 frames)               & 75.1    &   80.2    \\
        Sampling (300 frames)               & 73.2    &   79.0    \\
        Natural speed with attention mask     & \textbf{76.3}    &  \textbf{83.8}   \\
        \bottomrule
        \end{tabular} 
        }

\end{table}

\section{Learned Attention Maps}
\begin{figure*}[h]
\centering
    \includegraphics[width=9.5cm]{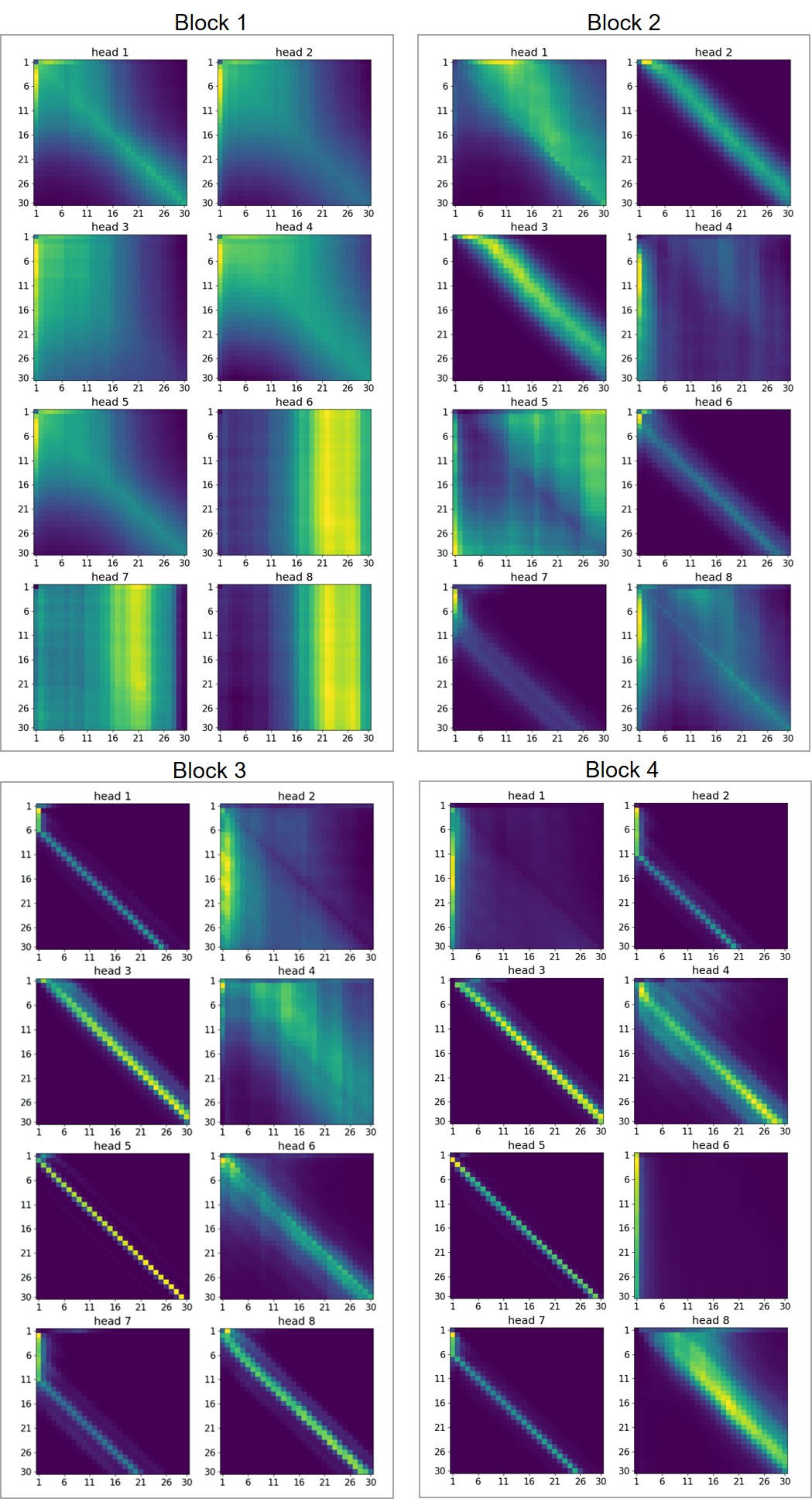}
    \caption{Learned temporal attention maps averaged over 300 evaluation sequences. Yellow color indicates a large value}
    \label{fig:supple_temporal_attention_maps}
\end{figure*}
\begin{figure*}[h]
\centering
    \includegraphics[width=9.5cm]{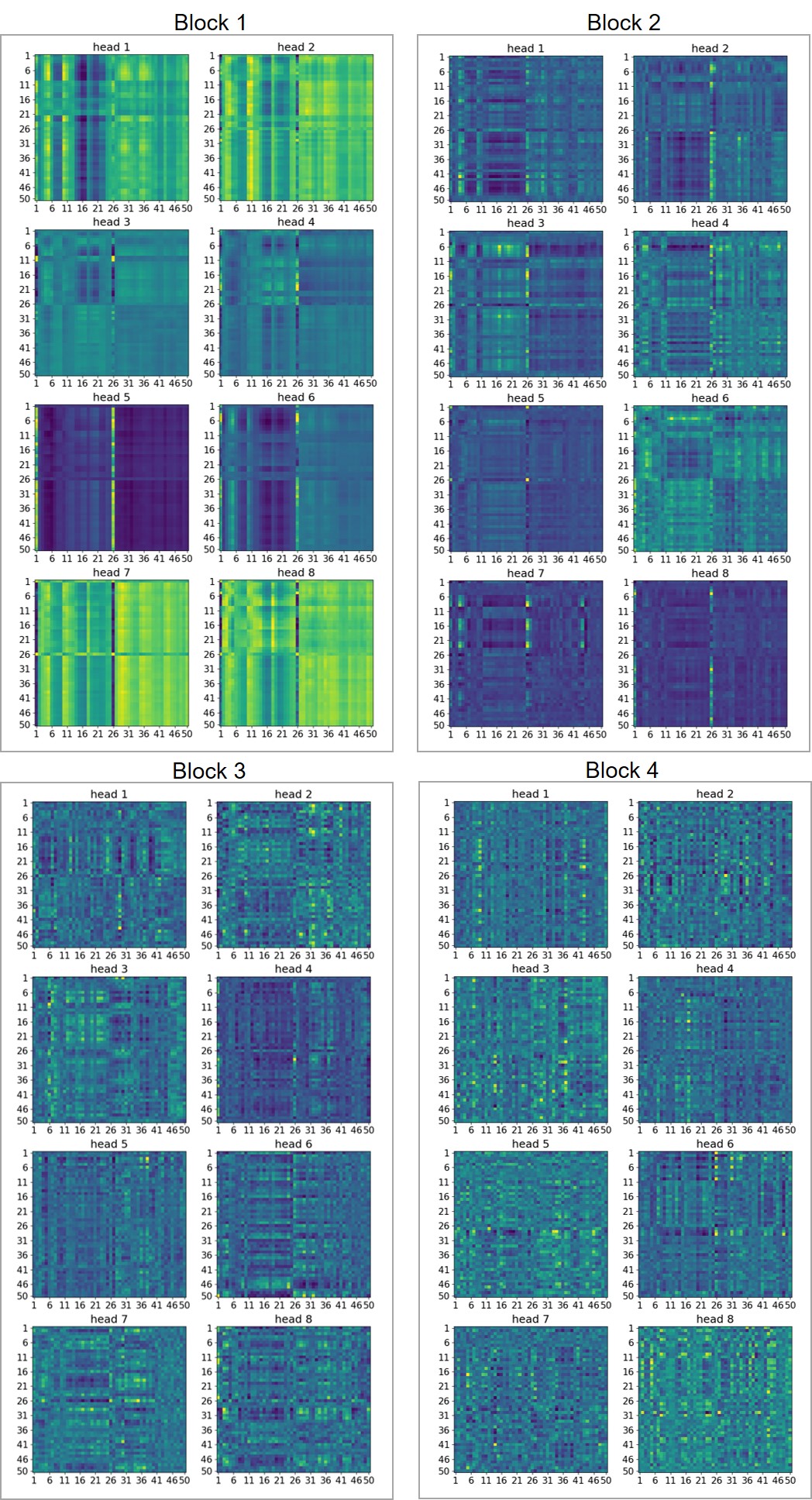}
    \caption{Learned spatial attention maps averaged over 300 evaluation sequences. Yellow color indicates a large value}
    \label{fig:supple_spatial_attention_maps}
\end{figure*}
Fig~\ref{fig:supple_temporal_attention_maps}. and Fig.~\ref{fig:supple_spatial_attention_maps} are supplementary figures for Fig. 4 and Fig. 5 in the main manuscript, respectively. 
In Fig.~\ref{fig:supple_temporal_attention_maps}, we illustrates the average of learned temporal attention maps for every head in each GL-Transformer block.
Likewise, Fig.~\ref{fig:supple_spatial_attention_maps} shows the average of learned spatial attention maps for every head in each GL-Transformer block.
The attention maps are averaged over 300 motion sequences of the evaluation data.

\section{Additional Study for Corrupted Input Sequence}
Some of the previous works~\cite{zheng2018unsupervised,su2020predict,cheng2021hierarchical} in unsupervised skeleton-based action learning perform pretraining tasks with randomly corrupted input sequences.
LongT GAN~\cite{zheng2018unsupervised} and P\&C~\cite{su2020predict} reconstruct the corrupted joints, and H-transformer~\cite{cheng2021hierarchical} predicts the instantaneous velocity of the joints for the masked (i.e. corrupted) frames following the scheme in \textit{masked token prediction} which is widely used in the transformer-based pretraining methods.
We conduct additional experiments with randomly corrupted input sequences to see their effects on the proposed method.
We randomly set the joint coordinates of the motion sequences to (0, 0, 0) and pretrain the proposed model with the corrupted sequences.
Indeed, the accuracy gradually decreases while the proportion of the corrupted joint increases, as presented in Table~\ref{table:supple_corrupt}.
Because our pretraining strategy is not to reconstruct the corrupted joints but to predict the displacement of all joints, the corrupted sequence rather causes the loss of information of the training data, resulting in lower performance.


\begin{table}[h]
\caption{Experimental results when the proposed model is pretrained on motion sequences with randomly corrupted joints.
The experiments follow the linear evaluation protocol on NTU-60 dataset.}
\label{table:supple_corrupt}
\centering
    \resizebox{0.43\textwidth}{!}{
    \begin{tabular}{l|CC}
    \toprule
    \multirow{2}{*}{\begin{tabular}[l]{@{}l@{}}Corrupted  joint\\ proportion\end{tabular}} & \multicolumn{2}{c}{Accuracy(\%)} \\
                                                                                       & xsub         & xview         \\ \midrule
    0\%     &   \textbf{76.3}    &   \textbf{83.8}    \\
    5\%     &   75.7    &   82.5    \\
    10\%    &   74.2    &   81.2    \\
    15\%    &   73.7    &   79.9    \\
    20\%    &   72.4    &   79.0    \\

    \bottomrule
    \end{tabular} 
    }
\end{table}


\end{document}